\documentclass[lettersize,journal]{IEEEtran}
\usepackage{amsmath,amsfonts}
\usepackage[linesnumbered,ruled,vlined]{algorithm2e}
\usepackage{array}
\usepackage{textcomp}
\usepackage{stfloats}
\usepackage{url}
\usepackage{verbatim}
\usepackage{graphicx}
\usepackage{cite}
\usepackage{amssymb}
\usepackage{mathtools}
\usepackage{amsthm}
\usepackage{enumitem}
\usepackage{microtype}
\usepackage{graphicx}
\usepackage{subcaption}
\usepackage{booktabs} 
\usepackage{tcolorbox}
\usepackage{balance}

\usepackage{multirow}
\usepackage{color}
\usepackage{bm}
\usepackage{newfloat}
\usepackage{listings}
\tcbuselibrary{listings, breakable}

\begin{document}

\title{Variational Cognitive Memory Abstraction for Long-Horizon Vision-Language-Action Models%
\thanks{Corresponding author: Chenwei~Tang}
\thanks{X. Liu, C. Tang, C. Zhang, Q Zeng, W Ju, and J. Lv are with the College of Computer Science, Sichuan University, and Engineering Research Center of Machine Learning and Industry Intelligence, Ministry of Education, Chengdu 610065, PR China (e-mail: xinyuliu0804@stu.scu.edu.cn, \{tangchenwei, lvjiancheng\}@scu.edu.cn).}
\thanks{D. Tian, R. Han and L. Zhen are with the Institute of High Performance Computing, Agency for Science, Technology and Research (A*STAR), Singapore 138632, Singapore}
}

\author{Xinyu~Liu,
        Daiying~Tian,
        Chenwei~Tang,
        Ruyang~Han, 
        Chengyang~Zhang, 
        Qingyu~Zeng, 
        Wei~Ju,
        Liangli~Zhen,
        and~Jiancheng~Lv}

\maketitle
\begin{abstract}
Vision-Language-Action (VLA) models have shown remarkable success in robotic manipulation, yet their ability to execute long-horizon tasks remains limited by insufficient long-term memory modeling. Existing methods either rely on implicit temporal representations within policy networks, where historical information can be gradually diluted, or maintain explicit memory buffers that suffer from redundancy, noise accumulation, and inefficient retrieval. In this paper, we propose \textbf{CogMemory}, a novel variational cognitive memory framework that enables VLA systems to selectively preserve and reuse task-relevant historical information. Our approach encodes multimodal trajectories into compact latent memory representations through a selective state-space memory encoder, multi-slot temporal abstraction, and variational information compression. During action generation, dynamic cross-attention retrieval with gated fusion adaptively incorporates relevant memories into policy prediction. With merely 3.45M additional parameters, \textbf{CogMemory} stays lightweight and seamlessly fits both early‑ and intermediate‑fusion VLA backbones. Extensive experiments on long-horizon robotic manipulation benchmarks demonstrate consistent improvements across diverse VLA backbones.
\end{abstract}

\begin{figure}[t]
\centering
\includegraphics[width=\columnwidth]{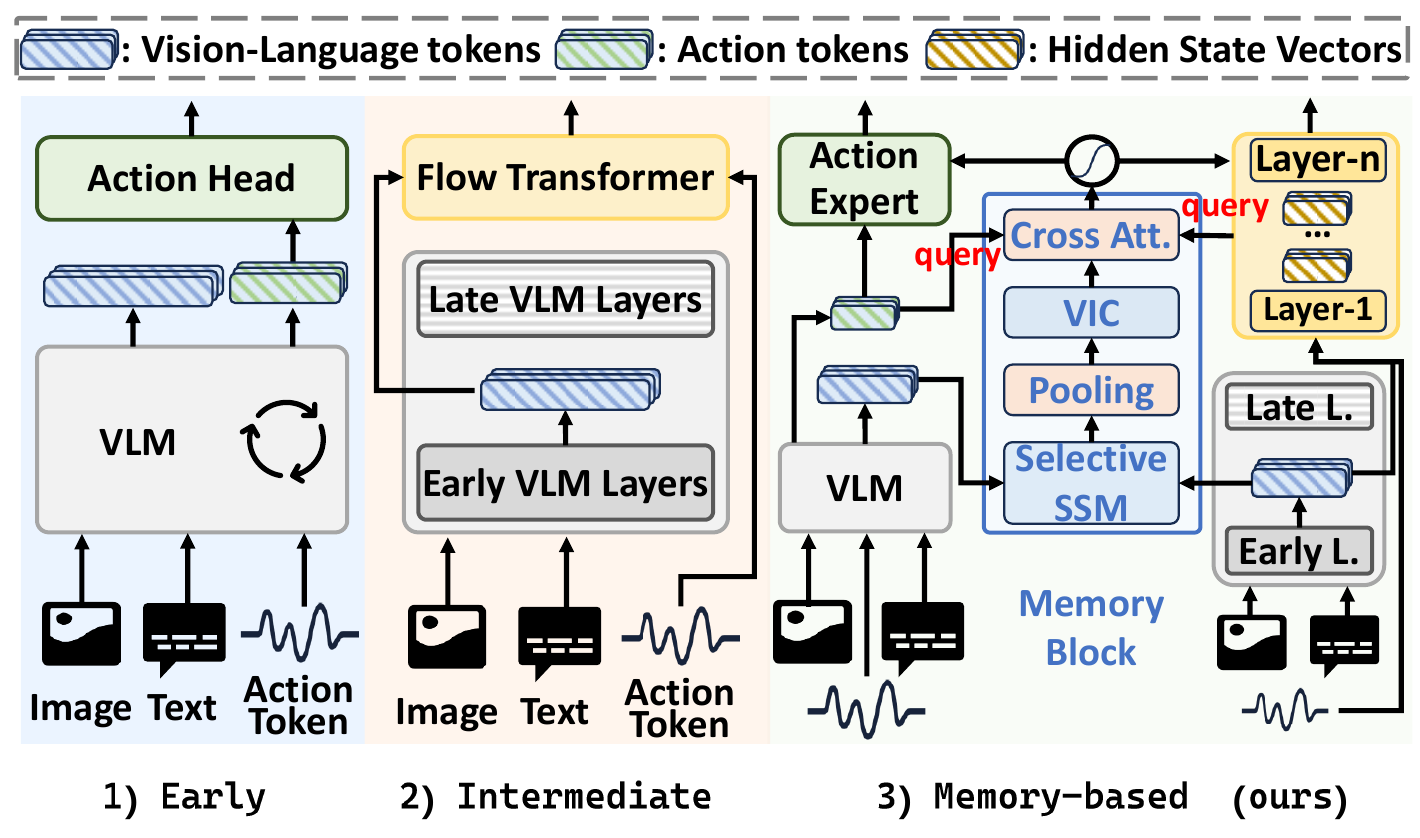}
\caption{Early-fusion architectures jointly model all modalities but lack explicit long-range history retention, while intermediate-fusion approaches rely on implicit temporal aggregation within policy networks, leaving them vulnerable to noise and decay of early signals. Our proposed \textbf{CogMemory} abstracts long-horizon history into compact latent slots via selective state-space modeling (SSM) and variational information compression (VIC), and enables the policy to dynamically query task-relevant memory via cross-attention and gated fusion for action generation. This lightweight module can be seamlessly integrated into both early-fusion and intermediate-fusion VLA models.}
\label{fig:motivation}
\end{figure}

\section{Introduction}
\label{sec:introduction}
Recent advances in Vision-Language-Action (VLA) models have significantly improved robotic manipulation through large-scale multimodal pretraining  \cite{zhang2025pure,ma2026survey,sapkota2025vision}. Existing VLA systems broadly follow two predominant fusion paradigms, as illustrated in Figure \ref{fig:motivation}. \textit{Early-fusion architectures} \cite{zitkovich2023rt,kim2024openvla,mees2024octo} jointly model vision, language, and actions within unified transformers, yet they lack explicit mechanisms for retaining long-range history—task-relevant cues from early steps are progressively diluted as interaction horizons grow, causing severe performance degradation in multi-stage tasks. \textit{Intermediate-fusion approaches}, e.g., diffusion \cite{ho2020denoising} and flow-based \cite{lipman2022flow} policies \cite{black2024pi_0,li2024cogact,reuss2025flower}, integrate VLM representations into continuous action generators, but they still rely on implicit temporal modeling within policy networks, where historical information is only passively aggregated through sequential updates. This implicit aggregation remains vulnerable to noise, redundancy, and the gradual decay of early critical signals, making it difficult to preserve long-term dependencies or maintain temporally consistent decisions across extended executions. To overcome these inherent limitations, recent works have introduced explicit memory mechanisms into VLA systems \cite{wang2021structured,li2025towards,shi2025memoryvla}, typically via external memory banks, recurrent states, or historical token caches retrieved during inference. While these methods improve temporal reasoning, they introduce three critical shortcomings, i.e., i) \textit{\textbf{Redundant Retrieval}}, direct storage of historical observations or intermediate features incurs severe redundancy and retrieval inefficiency as horizons scale, ii) \textit{\textbf{Noise Susceptibility}}, existing memory representations lack explicit abstraction, making them susceptible to task-irrelevant or noisy history, and iii) \textit{\textbf{Architectural Coupling}}, many such designs are tightly coupled to specific policy backbones, limiting their generality. These drawbacks motivate a lightweight, compact, and backbone-agnostic memory framework that can efficiently abstract long-term task-relevant information without compromising inference speed or architectural flexibility.

To address these challenges, we propose \textbf{CogMemory}, a lightweight variational memory abstraction framework tailored for long-horizon VLA models. Unlike conventional memory banks or token caches, \textbf{CogMemory} directly compresses extended histories into compact latent slots through two core mechanisms, i.e., a selective State-Space Model (SSM) that efficiently accumulates long-range temporal information and Variational Information Compression (VIC) that distills task-relevant signals while discarding redundancy. This design directly mitigates \textit{\textbf{redundant retrieval}}. Instead of storing raw observations or intermediate features, we maintain only a handful of structured latent slots, drastically reducing storage and retrieval costs. The SSM’s selective gating and the VIC objective jointly filter out noisy, task-irrelevant history, thereby overcoming \textit{\textbf{noise susceptibility}} and ensuring that only temporally coherent and informative cues are preserved. During action generation, a dynamic cross-attention mechanism coupled with gated fusion allows the policy to query these memory slots on demand, selectively incorporating relevant context while ignoring irrelevant distractions. This further reinforces robustness against noise. Moreover, \textbf{CogMemory} is designed as a plug-and-play module that integrates seamlessly into both early-fusion and intermediate-fusion VLA backbones, as shown in Figure \ref{fig:motivation}, effectively resolving \textit{\textbf{architectural coupling}} without requiring invasive modifications to the base policy. We evaluate \textbf{CogMemory} on representative long-horizon robotic manipulation benchmarks across diverse task settings. Experimental results demonstrate that explicit cognitive memory abstraction consistently improves long-horizon manipulation performance across diverse VLA backbones. These results highlight the effectiveness of compact explicit memory abstraction for improving long-horizon robotic decision making. Our contributions are summarized as follows:

\begin{itemize}[leftmargin=*, nosep, topsep=0pt]
    \item We propose \textbf{CogMemory}, a lightweight variational memory abstraction framework that compresses long-horizon interaction histories into compact latent slots via SSM for efficient temporal accumulation and VIC for redundancy filtering, enabling selective preservation of task-relevant cues while discarding irrelevant history.  
    \item We design a plug-and-play memory retrieval mechanism based on dynamic cross-attention and gated fusion, which allows the policy to query memory slots on demand and selectively incorporate historical context. This module is backbone-agnostic and seamlessly integrates into both early-fusion and intermediate-fusion VLA architectures.
    \item Extensive experiments demonstrate that \textbf{CogMemory} consistently and significantly improves performance across diverse VLA backbones, validating the effectiveness, generality, and lightweight nature of our approach for long-horizon sequential decision-making.
\end{itemize}

\section{Related Work}
\label{sec:related work}
VLA models have recently advanced general robotic manipulation by combining large vision-language models \cite{radford2021learning,liu2023visual,steiner2024paligemma} with large-scale robot datasets \cite{o2024open,ebert2021bridge,walke2023bridgedata}. Existing VLA approaches formulate robot control as multimodal sequence modeling, where visual observations, language instructions, and robot states are jointly encoded for action prediction. Representative methods like RT-series models \cite{brohan2022rt,zitkovich2023rt}, OpenVLA \cite{kim2024openvla}, Octo \cite{mees2024octo}, $\pi_0$ \cite{black2024pi_0}, CogACT \cite{li2024cogact}, and FLOWER \cite{reuss2025flower}, have explored different architectures for representation learning and continuous robot action generation. Despite their strong performance on individual manipulation tasks, long-horizon robotic control remains challenging due to the difficulty of preserving and exploiting relevant historical information over extended interactions. Recent works have introduced explicit memory mechanisms into robotic policies. RoboFlamingo \cite{li2024vision} propagates compressed vision-language representations through recurrent states, while TraceVLA \cite{zheng2025tracevla} and Interleave-VLA \cite{fan2025interleave} encode interaction histories into visual or multimodal sequences. UniVLA \cite{bu2025univla} incorporates previous actions for temporal reasoning, and MemoryVLA \cite{shi2025memoryvla} introduces an external memory retrieval mechanism for cognition-guided action generation. However, existing approaches either rely on implicit temporal accumulation or directly cache historical features, which may introduce redundancy and limit scalability for long-horizon manipulation. In contrast, \textbf{CogMemory} explicitly abstracts interaction histories into compact variational memory representations and adaptively retrieves task-relevant information for long-horizon decision making.

\begin{figure*}[t]
\centering
\includegraphics[width=0.95\textwidth]{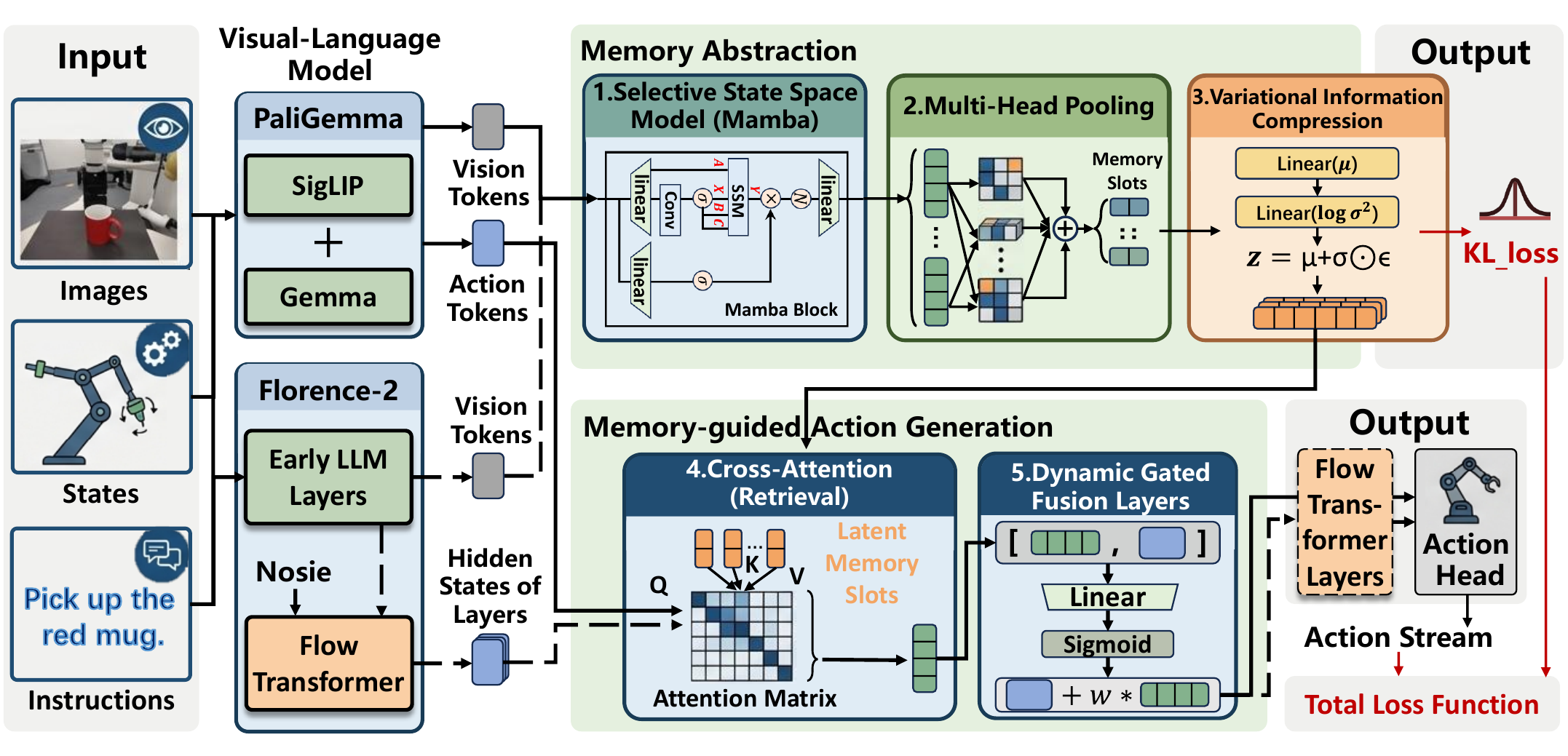}
\caption{Overview of \textbf{CogMemory}. Given visual observations, robot states, and language instructions, the VLA backbone first produces multimodal representations. \textbf{CogMemory} then constructs compact cognitive memory through three stages: (1) selective state-space modeling (SSM) for long-range temporal encoding, (2) multi-slot temporal pooling for memory abstraction, and (3) variational information compression (VIC) for learning informative latent memory distributions. During action generation, the action decoder queries task-relevant cognitive memory through cross-attention, followed by dynamic gated fusion to adaptively integrate queried memory into action prediction. The framework is compatible with both early-fusion (e.g., $\pi_0$-style architectures) and intermediate-fusion VLA models (e.g., FLOWER-style architectures).}
\label{fig:framework}
\end{figure*}

\section{Method}
\label{sec:method}

As illustrated in Figure~\ref{fig:framework}, given multimodal representations extracted from a VLA backbone, \textbf{CogMemory} introduces a two-stage cognitive memory pipeline: \textit{\textbf{Memory Abstraction}} and \textit{\textbf{Memory-guided Action Generation}}. The abstraction stage sequentially compresses long-horizon interaction histories into compact latent memory slots through three core operations, i.e., i) selective state-space modeling (SSM) for long-range temporal encoding, ii) multi-slot temporal pooling for structured memory consolidation, and iii) variational information compression (VIC) for learning informative and noise-robust latent memory distributions. The action generation stage then retrieves task-relevant memory via cross-attention and adaptively fuses it with current representations through gated fusion, enabling temporally coherent and context-aware decision making.

Formally, let
$
X = [x_1, x_2, \dots, x_L] \in \mathbb{R}^{L \times D}
$
denote the multimodal representations extracted from the VLA backbone, where $L$ is the history length and $D$ is the feature dimension. Depending on the backbone architecture, X may correspond to multimodal token representations from VLM outputs (e.g., $\pi_0$-style architectures) or intermediate representations extracted from early fusion layers (e.g., FLOWER-style architectures). \textbf{CogMemory} transforms $X$ into a structured latent memory representation
$
M \in \mathbb{R}^{K \times D},
$
where $K \ll L$ denotes the number of memory slots. During policy inference, task-relevant historical information is dynamically retrieved from M and adaptively integrated into action generation for long-horizon decision making.

\subsection{Long-Horizon Memory Modeling in VLA}
\label{sec: method-1}
At each timestep $t$ of long-horizon robotic manipulation under the VLA framework, the robot receives multimodal observations consisting of visual inputs $o_t$, robot states $s_t$, and a language instruction $l$. The policy aims to predict actions conditioned on the accumulated interaction history:
\begin{equation}
a_t \sim \pi_\theta(a_t \mid o_{\leq t}, s_{\leq t}, l),
\end{equation}
where $o_{\leq t}=\{o_1,\dots,o_t\}$ and $s_{\leq t}=\{s_1,\dots,s_t\}$ denote historical observations and robot states up to timestep $t$.

Existing VLA systems typically encode temporal dependencies implicitly through sequential representations within policy networks. While this suffices for short-horizon tasks, implicit temporal modeling struggles to preserve long-range task-relevant dependencies over extended interaction horizons. Taking transformer-based VLA policies as a representative example, temporal reasoning is formulated as:
\begin{equation}
h_t
=
\sum_{i=1}^{t}
\alpha_{t,i} W_v x_i,
\qquad
\alpha_{t,i}
=
\frac{
\exp(q_t^\top k_i)
}{
\sum_{j=1}^{t}\exp(q_t^\top k_j)
},
\end{equation}
where historical representations are aggregated through attention weights $\alpha_{t,i}$. As interaction horizons grow, the self-attention mechanism must distribute its limited capacity over increasingly lengthy histories, making it difficult to consistently attend to and preserve task-relevant cues from distant timesteps. Consequently, early observations become progressively diluted by subsequent activations, resulting in unstable temporal reasoning and degraded retention of long-range dependencies in multi-stage manipulation tasks.

To alleviate this issue, memory-based VLA approaches explicitly maintain historical representations through external memory mechanisms
$
\mathcal{M}
=
\{m_1,m_2,\dots,m_T\},
$
which are retrieved during policy inference. While explicit retrieval improves temporal reasoning, directly caching historical features introduces two fundamental limitations. First, memory complexity scales linearly with interaction horizon
$
|\mathcal{M}| = \mathcal{O}(T),
$
leading to prohibitive storage and retrieval costs over long-horizon execution. Second, raw historical hidden states contain significant task-irrelevant information and observation noise. Since retrieval operates directly on uncompressed historical representations without any abstraction, the retrieved memories are prone to redundancy and noise contamination, undermining the reliability of downstream action prediction. To overcome these limitations, we propose \textbf{CogMemory}, a lightweight cognitive memory abstraction framework that compresses long-horizon interaction histories into compact latent representations rather than naively caching raw historical features. Through selective temporal encoding, variational memory compression, and adaptive retrieval, \textbf{CogMemory} enables efficient long-term memory modeling for VLA systems while remaining agnostic to the underlying backbone architecture.

\subsection{Memory Abstraction}
\label{sec:method-3}
Given multimodal representations $X$, we construct compact temporal memory abstractions through a progressive three-stages: i) selective SSM for long-range temporal encoding, ii) multi-head pooling for structured slot consolidation, and iii) VIC for learning informative and noise-robust latent memory distributions. These components work in concert to progressively distill raw interaction histories into compact, structured, and task-relevant cognitive memory representations.

\subsubsection{Selective State-Space Temporal Encoding.}
We first employ a selective state-space model (SSM)\cite{gu2023mamba,dao2024transformers} to efficiently encode long-range temporal dependencies from interaction histories: 
\begin{equation}
H=\mathrm{SSM}(X),
\end{equation}
where
$
H=[h_1,h_2,\dots,h_L]\in\mathbb{R}^{L\times D}.
$
Unlike attention-based temporal aggregation, selective SSMs maintain historical information through recurrent state propagation:
\begin{equation}
h_t=A_t h_{t-1}+B_t x_t,
\end{equation}
where $A_t$ and $B_t$ denote input-dependent state transition matrices. Unrolling the recurrence yields:
\begin{equation}
h_t
=
\left(
\prod_{i=1}^{t}A_i
\right)h_0
+
\sum_{j=1}^{t}
\left(
\prod_{i=j+1}^{t}A_i
\right)
B_jx_j.
\end{equation}
This formulation enables efficient temporal propagation through structured recurrent state evolution. Compared with conventional recurrent modules, selective SSMs provide efficient history aggregation through structured state dynamics. The resulting representations provide a stable basis for subsequent memory abstraction over long interaction horizons. Moreover, the input-dependent gating of the selective SSM dynamically modulates state transitions conditioned on current observations, allowing the model to emphasize task-relevant temporal cues while attenuating redundant historical information. This yields temporally structured representations that are readily amenable to the subsequent pooling and compression stages.

\subsubsection{Multi-Head Pooling.}
To construct structured long-term memory representations, we summarize temporal features into multiple latent memory slots. Specifically, we first compute slot assignment logits:
\begin{equation}
A=HW_p,
\end{equation}
where
$
A\in\mathbb{R}^{L\times K},
$
$
W_p\in\mathbb{R}^{D\times K},
$
and $K$ denotes the number of memory slots. The assignment weights are normalized across the temporal dimension to aggregate contributions from different timesteps:
\begin{equation}
\alpha_{t,k}
=
\frac{
\exp(A_{t,k})
}{
\sum_{t'=1}^{L}\exp(A_{t',k})
}.
\end{equation}

Each memory slot $m_k$ is then constructed through weighted temporal aggregation:
\begin{equation}
m_k = \sum_{t=1}^{L} \alpha_{t,k}h_t, \quad M_{\mathrm{temp}} = [m_1,\dots,m_K].
\end{equation}
Unlike single-vector memory compression, multi-slot abstraction $M_{temp} \in \mathbb{R}^{K\times D}$ organizes long-horizon interaction history into multiple structured semantic components. Each memory slot aggregates complementary temporal patterns, allowing different slots to specialize in different interaction stages, subgoals, or manipulation patterns, reducing redundancy between temporally distant observations while improving memory compactness.

\subsubsection{Variational Information Compression.}
Although multi-slot abstraction produces structured temporal memory representations, the resulting features may still contain redundant historical information and observation noise. To further encourage compact and informative memory representations, we introduce a VIC to model memory representations as latent distributions rather than deterministic feature vectors. Specifically, each memory slot is projected into a latent Gaussian distribution:
\begin{equation}
q(z_k|m_k)
=
\mathcal{N}(\mu_k,\sigma_k^2),
\end{equation}
where $\mu_k=f_\mu(m_k)$ and $\log\sigma_k^2=f_\sigma(m_k)$. We then sample latent memory representations by reparameterization trick:
\begin{equation}
z_k
=
\mu_k+\sigma_k\odot\epsilon,
\qquad
\epsilon\sim\mathcal{N}(0,I).
\end{equation}

Following the VIC principle \cite{tishby2000information,tishby2015deep}, VIC encourages the latent memory representation to retain task-relevant information while reducing unnecessary historical details:
\begin{equation}
\max I(Z;y)-\beta I(Z;x),
\end{equation}
where $x$ denotes historical observations and $y$ represents task-relevant latent semantics.

Compared with deterministic memory encoding, distributional memory modeling encourages temporally similar interaction histories to share smoother latent representations, improving robustness against noisy observations and temporal perturbations. Furthermore, the KL regularization:
\begin{equation}
\mathcal{L}_{\mathrm{KL}}
=
\sum_{k=1}^{K}
D_{\mathrm{KL}}
\big(
q(z_k|m_k)\,\|\,p(z_k)
\big),
\quad
p(z_k)=\mathcal{N}(0,I).
\end{equation}
constrains the latent memory space toward a compact prior distribution, suppressing redundant trajectory-specific details and improving memory consistency across long-horizon interaction trajectories.

Finally, the sampled latent variables are projected into the final cognitive memory representation:
\begin{equation}
M=f_{\mathrm{proj}}(Z) \in\mathbb{R}^{K\times D},
\end{equation}
where $Z=[z_1,\dots,z_K]$ denotes the sampled latent memory variables. The variational formulation further encourages smooth latent memory representations under temporally correlated interaction trajectories.

\begin{table*}[t]
\centering
\setlength{\tabcolsep}{11pt}
\small
\begin{tabular}{lcccccc}
\toprule
\textbf{Method} & \textbf{Spatial$\uparrow$} & \textbf{Object$\uparrow$} & \textbf{Goal$\uparrow$} & \textbf{Long-10$\uparrow$} & \textbf{Long-90$\uparrow$} & \textbf{Avg.$\uparrow$} \\ 
\midrule
UniACT  \cite{zheng2025universal} & 77.0 & 87.0 & 77.0 & 70.0 & 73.0 & 76.8 \\
SpatialVLA  \cite{qu2025spatialvla} & 88.2 & 89.9 & 78.6 & 55.5 & 46.2 & 71.7 \\
TraceVLA  \cite{zheng2025tracevla} & 84.6 & 85.2 & 75.1 & 54.1 & --   & 74.8 \\
CoT-VLA  \cite{zhao2025cot} & 87.5 & 91.6 & 87.6 & 69.0 & --   & 83.9 \\
$\pi_0$-FAST  \cite{pertsch2025fast} & 96.4 & 96.8 & 88.6 & 60.2 & 83.1 & 85.0 \\
CronusVLA  \cite{li2025cronusvla} & 90.1 & 94.7 & 91.3 & 68.7 & --   & 86.2 \\
TriVLA  \cite{liu2025trivla} & 91.2 & 93.8 & 89.8 & 73.2 & --   & 87.0 \\
4D-VLA  \cite{zhang20264d} & 93.8 & 92.8 & 95.6 & 86.5 & --   & 92.2 \\
$\pi_{0.5}$  \cite{intelligence2025pi_} & 98.0 & 97.8 &95.6 & 85.8 & 96.0 & 94.6 \\
UniVLA \cite{bu2025univla} & 96.5 & 96.8 & 95.6 & 92.0 & 95.2 & 95.2\\
MemoryVLA  \cite{shi2025memoryvla} & \textbf{98.4} & 98.4 & 96.4 & 93.4 & 95.6 & 96.5\\
FLOWER  \cite{reuss2025flower} & 97.5 & \textbf{99.1} & 96.1 & 94.9 & 94.7 & 96.5\\
\midrule
\textbf{CogMemory (Ours)} & 98.2 & 98.8 & \textbf{97.2} & \textbf{97.1} & \textbf{96.9} & \textbf{97.6} \\
\bottomrule
\end{tabular}
\caption{Comparison of the task success rates (\%) on LIBERO using the Franka Panda robot across five evaluation suites. For methods that do not report Long-90 results, we follow the original reporting protocol and compute Avg. over the available suites.}
\label{tab:libero_result}
\end{table*}

\subsection{Memory-Guided Action Generation}
\label{sec:method-4}
Given the latent memory representation $M$, we adaptively integrate retrieved historical information into action generation through memory-conditioned fusion. This fusion mechanism is designed to seamlessly accommodate diverse VLA architectures. For early-fusion models such as $\pi_0$-style architectures, where actions are generated from multimodal prefix representations and action-related tokens, \textbf{CogMemory} applies memory-conditioned cross-attention directly to the action suffix representations: 
\begin{equation}
R
=
\mathrm{Softmax}
\left(
\frac{
Q_{\mathrm{suffix}}M^\top
}{
\sqrt{D}
}
\right)
M,
\end{equation}
where $Q_{\mathrm{suffix}}
\in
\mathbb{R}^{N\times D}$ denotes the action-generation suffix representations and $\mathrm{Softmax}$ is applied along the memory-slot dimension. The retrieved memory feature is then adaptively integrated into the action representation through gated residual fusion:
\begin{equation}
G
=
\sigma
\left(
W_g[Q_{\mathrm{suffix}};R]
\right),
\quad
\tilde{Q}_{\mathrm{suffix}}
=
Q_{\mathrm{suffix}}
+
G\odot R.
\end{equation}
where $[\cdot;\cdot]$ denotes feature concatenation. The fused representation $\tilde{Q}_{\mathrm{suffix}}$ is used for downstream action prediction. 

In contrast, intermediate-fusion architectures such as FLOWER-style models perform action generation through iterative hidden-state refinement in Flow Transformer layers. In this setting, \textbf{CogMemory} can be integrated into intermediate hidden representations during Flow Transformer computation:
\begin{equation}
R^{(l)}
=
\mathrm{Softmax}
\left(
\frac{
Q^{(l)}M^\top
}{
\sqrt{D}
}
\right)
M,
\end{equation}
where $Q^{(l)}\in\mathbb{R}^{N\times D}$ denotes the hidden states at the $l$-th Flow Transformer layer. The retrieved memory is then fused through adaptive gating:
\begin{equation}
G^{(l)}
=
\sigma
\left(
W_g[Q^{(l)};R^{(l)}]
\right),
\quad
\tilde{Q}^{(l)}
=
Q^{(l)}
+
G^{(l)}\odot R^{(l)}.
\end{equation}

By decoupling memory abstraction from backbone-specific policy computation, \textbf{CogMemory} provides a lightweight and versatile interface for incorporating long-horizon memory into heterogeneous VLA architectures. This module is jointly optimized with the underlying VLA policy in an end-to-end manner, enabling the memory representations to adapt to the policy's specific temporal reasoning demands. The overall training objective comprises the original training objective $\mathcal{L}_{\mathrm{action}}$ defined by each VLA backbone along with a variational regularization term that enforces compact and informative latent memory distributions:
\begin{equation}
\mathcal{L}
=
\mathcal{L}_{\mathrm{action}}
+
\lambda\mathcal{L}_{\mathrm{KL}},
\end{equation}
where $\lambda$ controls the strength of variational compression.

\section{Experiment}
\label{sec:experiment}
In this section, we conduct extensive experiments to evaluate the effectiveness of \textbf{CogMemory} for long-horizon robotic manipulation on two widely used long-horizon robotic manipulation benchmarks: LIBERO \cite{liu2023libero} and CALVIN \cite{mees2022calvin}. LIBERO contains 130 language-conditioned manipulation tasks organized into five suites covering spatial manipulation, object interaction, goal-conditioned control, and long-horizon task composition. CALVIN evaluates language-conditioned manipulation through sequential compositions of 34 manipulation primitives with over 380 language annotations across multiple environment splits. Together, these benchmarks provide complementary evaluations of long-horizon manipulation, task composition, and language-conditioned control. 
\begin{table*}[t]
\centering
\small
\begin{tabular}{l l l c c c c c c}
\toprule
\textbf{Setting} & \textbf{Method} & \textbf{Backbone} & \textbf{LH-1 $\uparrow$} & \textbf{LH-2 $\uparrow$} & \textbf{LH-3 $\uparrow$} & \textbf{LH-4 $\uparrow$} & \textbf{LH-5 $\uparrow$} & \textbf{Avg. Len. $\uparrow$} \\
\midrule
\multirow{7}{*}{D$\to$D} & $\pi_{0.5}$ \cite{intelligence2025pi_} & PaliGemma & 86.8 & 68.8 & 53.3 & 42.5 & 32.4 & 2.84 \\
 & \textbf{CogMemory (Ours)} & PaliGemma & \textbf{89.7} & \textbf{75.5} & \textbf{63.5} & \textbf{53.6} & \textbf{44.9} & \textbf{3.27} \\
\cline{2-9}
 & MDT \cite{reuss2024multimodal} & Diffusion & 93.7 & 84.5 & 74.1 & 64.4 & 55.6 & 3.72 \\
 & RoboUniView \cite{liu2024robouniview} & Cont & 96.2 & 88.8 & 77.6 & 66.6 & 56.3 & 3.85 \\
 & FLOWER \cite{reuss2025flower} & Florence-2-L & 97.4 & 92.4 & 86.9 & 81.3 & 74.9 & 4.35 \\
 & ThinkProprio \cite{wang2026think} & Florence-2-L & 96.9 & 89.8 & 83.6 & 80.5 & 72.7 & 4.23 \\
 & \textbf{CogMemory (Ours)} & Florence-2-L & \textbf{98.1} & \textbf{93.9} & \textbf{88.2} & \textbf{83.3} & \textbf{78.1} & \textbf{4.42} \\
\specialrule{1.0pt}{0pt}{0pt} 
\multirow{8}{*}{ABC$\to$D} & RoboFlamingo \cite{li2024vision} & ViT & 82.4 & 61.9 & 46.6 & 33.1 & 23.5 & 2.47 \\
 & $\pi_0$ \cite{black2024pi_0} & PaliGemma & 70.0 & 48.0 & 37.0 & 28.0 & 18.0 & 2.01 \\
 & $\pi_{0.5}$ \cite{intelligence2025pi_} & PaliGemma & 71.0 & 56.0 & 45.0 & 37.0 & 29.0 & 2.38 \\
 & \textbf{CogMemory (Ours)} & PaliGemma & \textbf{89.3} & \textbf{69.9} & \textbf{55.1} & \textbf{42.8} & \textbf{35.3} & \textbf{2.92} \\
 \cline{2-9}
 & OpenVLA \cite{kim2024openvla} & DINOv2+SigLIP & 91.3 & 77.8 & 62.0 & 52.1 & 43.5 & 3.27 \\
 & FLOWER \cite{reuss2025flower} & Florence-2-L & 99.3 & 96.0 & 90.3 & 82.3 & 75.5 & 4.44 \\
 & ThinkProprio \cite{wang2026think} & Florence-2-L & 97.7 & 96.1 & 92.2 & 86.7 & 82.1 & 4.55 \\
 & \textbf{CogMemory (Ours)} & Florence-2-L & \textbf{98.7} & \textbf{97.5} & \textbf{94.6} & \textbf{89.1} & \textbf{84.4} & \textbf{4.64} \\
\bottomrule
\end{tabular}
\caption{Results on CALVIN benchmark with the Franka Emika robot. We report success rates (\%) for different task-chain lengths and the average completed chain length (Avg. Len.) under two evaluation settings: D$\to$D and ABC$\to$D.}
\label{tab:calvin_result}
\end{table*}

Our evaluation is designed to address the following three Research Questions (RQs): 
RQ1:
Can \textbf{CogMemory} improve long-horizon manipulation performance while maintaining overall task execution capability?
RQ2:
Can \textbf{CogMemory} improve sequential task completion over extended horizons and maintain effectiveness under unseen environment transfer?
RQ3:
How do different design components and memory slot capacities affect the performance of \textbf{CogMemory}? All experiments are conducted on 8 NVIDIA L20X GPUs and trained for 30k optimization steps using the AdamW optimizer \cite{zhou2024towards} with a global batch size of 256. We evaluate \textbf{CogMemory} on two representative VLA backbones covering different fusion paradigms: the 3B-parameter PaliGemma backbone \cite{beyer2024paligemma} (\(\pi_{0.5}\)-style) for early-fusion architectures and the 0.77B-parameter Florence-2-L backbone \cite{xiao2024florence} (FLOWER-style) for intermediate-fusion architectures. All models are trained following the standard benchmark protocols with trajectories sampled at 10 Hz. The proposed memory module introduces 3.45M additional parameters while enabling end-to-end optimization with the underlying VLA policy. Additional experiment details, hyperparameters, qualitative analysis, and efficiency comparisons are provided in \textbf{Appendix}.

\subsection{Long-Horizon Performance \& Preservation (RQ1)}
\label{sec:libero}
As shown in Table~\ref{tab:libero_result}, we compare \textbf{CogMemory} with recent VLA baselines and memory-augmented robotic policies on LIBERO benchmark \cite{liu2023libero} across five suites using a Franka Panda robot. Following prior works \cite{kim2024openvla,reuss2025flower}, each task is trained with 50 demonstrations for 30k steps. Validation is performed every 1k steps, and we report the results from the best validation checkpoint.  compares CogMemory with recent VLA baselines and memory-augmented policies. \textbf{CogMemory} achieves the best overall performance with an average success rate of 97.6\%, surpassing all prior approaches. This gain is most pronounced on the long-horizon suites, where CogMemory attains 97.1\% on Long-10 and 96.9\% on Long-90—substantially outperforming prior memory-based methods. These methods rely on directly caching raw historical features, making them vulnerable to redundant retrieval and noise susceptibility as horizons extend. In contrast, CogMemory's SSM and VIC mechanisms distill long histories into compact, task-relevant latent slots, effectively mitigating the progressive dilution of early signals. Meanwhile, strong performance on Spatial, Object, and Goal confirms that our approach preserves general task capability while enhancing long-horizon reasoning. 

\subsection{Sequential Capability \& Transferability (RQ2)}
\label{sec:calvin}
We further evaluate \textbf{CogMemory} on the CALVIN benchmark \cite{mees2022calvin} under both D$\rightarrow$D and ABC$\rightarrow$D settings using the standard long-horizon language-conditioned manipulation protocol. All models are fine-tuned on the training split for 30k steps. During evaluation, policies are required to follow free-form language instructions and sequentially complete 5 manipulation tasks over 1,000 instruction chains. Tables~\ref{tab:calvin_result} summarize the results across two backbone architectures: PaliGemma-based $\pi_{0.5}$ (early-fusion) and Florence-2-L-based (intermediate-fusion) FLOWER, demonstrating that \textbf{CogMemory} consistently improves performance regardless of the underlying VLA design. On D$\rightarrow$D, \textbf{CogMemory} boosts Avg. Len. from 2.84 to 3.27 on PaliGemma and from 4.35 to 4.42 on Florence-2-L. The gains grow with chain length (e.g., LH-5 improves from 32.4 to 44.9 and from 74.9 to 78.1, respectively), suggesting that the compact latent slots produced by SSM and VIC become increasingly advantageous as history accumulates, precisely where redundant retrieval and noise susceptibility would otherwise impair performance. Under the more challenging ABC$\rightarrow$D setting, \textbf{CogMemory} maintains consistent gains across all lengths, improving Avg. Len. from 2.38 to 2.92 on PaliGemma and from 4.44 to 4.64 on Florence-2-L. These results indicate that the task-relevant abstraction learned by \textbf{CogMemory} generalizes effectively to unseen environments by filtering out environment-specific noise while preserving semantically meaningful historical cues. Crucially, the improvements on both backbones verify the architectural decoupling of \textbf{CogMemory}, it integrates seamlessly into early-fusion and intermediate-fusion architectures without invasive modifications, while delivering reliable gains across both in-distribution and transferred long-horizon settings.
\begin{table}[t]
\centering
\setlength{\tabcolsep}{5pt}
\resizebox{\linewidth}{!}{
\small
\begin{tabular}{l l c c c c c c}
\toprule
\textbf{Variant} & \textbf{Method} & \textbf{L-1} & \textbf{L-2} & \textbf{L-3} & \textbf{L-4} & \textbf{L-5} & \textbf{Avg. Len.} \\
\midrule
\multirow{4}{*}{Comp.} & \textbf{Ours} & \textbf{89.7} & \textbf{75.5} & \textbf{63.5} & \textbf{53.6} & \textbf{44.9} & \textbf{3.27} \\
 & w/o Memory & 85.7 & 68.6 & 53.2 & 42.4 & 33.0 & 2.83 \\
 & w/o VIC & 87.3 & 67.7 & 50.2 & 39.4 & 30.3 & 2.75 \\
 & w/o Gate & 89.6 & 70.9 & 53.6 & 42.6 & 33.6 & 2.90 \\
\specialrule{\heavyrulewidth}{0pt}{0pt}
\multirow{3}{*}{\# Slots} & Slots=4 & 86.9 & 70.1 & 57.1 & 47.2 & 38.2 & 2.99 \\
 & Slots=8 & \textbf{89.7} & \textbf{75.5} & \textbf{63.5} & \textbf{53.6} & \textbf{44.9} & \textbf{3.27} \\
 & Slots=16 & 87.5 & 68.0 & 53.1 & 42.8 & 34.1 & 2.85 \\
\bottomrule
\end{tabular}}
\caption{Ablation studies on CALVIN D$\to$D using the PaliGemma-based backbone. We ablate (top) key components of CogMemory and (bottom) the number of memory slots. Success rates (\%)$\uparrow$ are reported at each chain length, along with the average completed chain length (Avg. Len.)$\uparrow$.}
\label{tab:ablation}
\end{table}

\subsection{Ablation Study (RQ3)}
\label{sec:ablation_study}
We conduct ablation studies on the CALVIN D$\rightarrow$D benchmark using the PaliGemma-based backbone to analyze the contribution of individual components and the effect of memory slot count in \textbf{CogMemory}. Following the same protocol as the main CALVIN experiments, all variants are fine-tuned on the official training split for 30k steps and evaluated on 1,000 instruction chains. We report success rates at each chain length along with Avg. Len.. Table~\ref{tab:ablation} presents the results. For component ablations, we consider three variants: ``w/o Memory'' removes the entire memory abstraction and retrieval module, reverting to the original backbone for action generation; ``w/o VIC'' replaces variational latent memory modeling with deterministic feature encoding, omitting distribution sampling and KL regularization; ``w/o Gate'' removes the adaptive gating mechanism and directly injects retrieved memory features without fusion control. Removing the memory module causes substantial degradation across all chain lengths, with the gap widening as the horizon grows (e.g., LH-5 drops by over 11\%), confirming that explicit memory abstraction is critical for preserving task-relevant historical information in sequential manipulation. Replacing VIC with deterministic encoding leads to the largest degradation on long-horizon chains, indicating that variational compression is essential for learning compact and noise-robust latent representations. Removing the gating mechanism also consistently reduces performance, suggesting that adaptive fusion—enabling dynamic regulation of memory utilization conditioned on current states—is important for effective integration of historical context. For the slot count ablation, we evaluate configurations with $K=4$, $8$, and $16$ memory slots. The best performance is achieved with $K=8$, suggesting that a moderate number of compact latent slots provides the optimal trade-off between memory compactness and long-horizon task execution capability.
\begin{figure}[t]
    \centering
    \includegraphics[width=\linewidth]{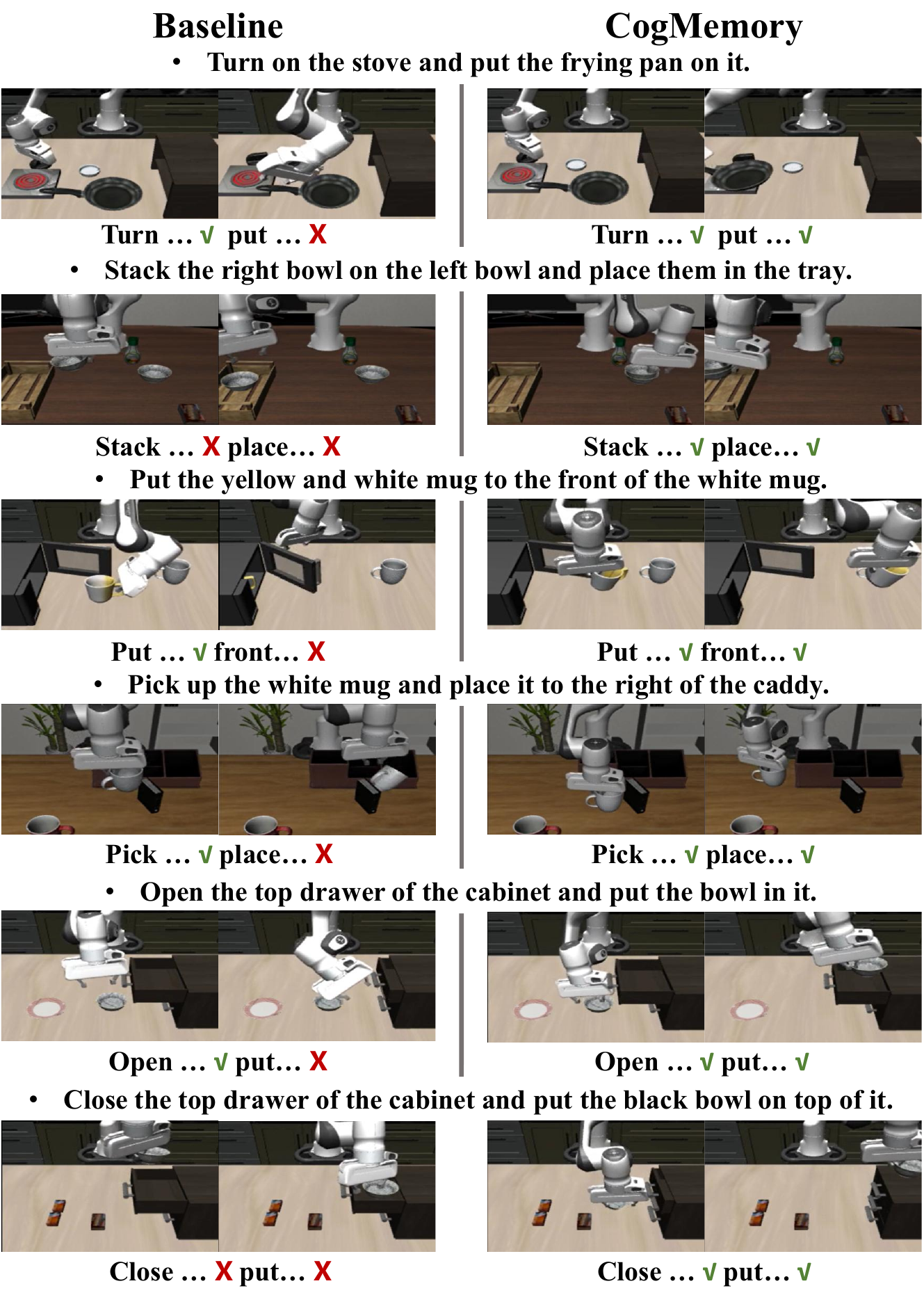}
    \caption{
    Qualitative comparison between the baseline policy and \textbf{CogMemory} on representative long-horizon manipulation tasks. Each row shows a sequential task requiring multiple operations, including stacking, turning, placing, picking, opening, and closing. Red and green markers indicate failure and successful completion, respectively.}
    \label{fig:qualitative}
\end{figure}

\subsection{Qualitative Analysis}
\label{sec:qualitative_analysis}
As shown in Figure~\ref{fig:qualitative}, to further understand how CogMemory improves long-horizon manipulation, we provide qualitative comparisons on representative sequential tasks. The baseline policy frequently fails in such scenarios, often completing early sub-tasks but failing on subsequent steps due to the gradual dilution of task-relevant historical context, a direct manifestation of the implicit temporal modeling limitation discussed earlier. In contrast, CogMemory consistently completes the full sequences, demonstrating that compact latent memory slots effectively preserve critical historical cues (e.g., ``stove is on'' or ``drawer is open'') across extended horizons. The selective retrieval and gated fusion mechanisms enable the policy to recall past states on demand, reducing error accumulation and improving robustness under sequential manipulation settings. These qualitative results complement our quantitative findings, further validating that explicit cognitive memory abstraction supports temporally coherent decision-making in long-horizon robotic tasks.

\section{Conclusion}
\label{conclusion}
In this paper, we proposed \textbf{CogMemory}, a lightweight memory abstraction framework that improves the long-horizon manipulation capability of VLA models through structured temporal abstraction, variational memory compression, and adaptive memory-guided action fusion. With only 3.45M additional parameters, \textbf{CogMemory} consistently improves performance across LIBERO and CALVIN benchmarks with different VLA architectures, demonstrating that compact and task-relevant memory representations can effectively enhance temporal reasoning and sequential task execution in robotic foundation models. Future work will investigate more scalable memory architectures for lifelong robot learning, continual adaptation, and open-world manipulation with increasingly diverse and complex experiences.

\bibliographystyle{ieeetr}
\bibliography{main}

\begin{thebibliography}{10}

\bibitem{zhang2025pure}
D.~Zhang, J.~Sun, C.~Hu, X.~Wu, Z.~Yuan, R.~Zhou, F.~Shen, and Q.~Zhou, ``Pure vision language action (vla) models: A comprehensive survey,'' {\em arXiv preprint arXiv:2509.19012}, 2025.

\bibitem{ma2026survey}
Y.~Ma, Z.~Song, Y.~Zhuang, J.~Hao, and I.~King, ``A survey on vision--language--action models for embodied ai,'' {\em IEEE Transactions on Neural Networks and Learning Systems}, 2026.

\bibitem{sapkota2025vision}
R.~Sapkota, Y.~Cao, K.~I. Roumeliotis, and M.~Karkee, ``Vision-language-action models: Concepts, progress, applications and challenges,'' {\em arXiv preprint arXiv:2505.04769}, 2025.

\bibitem{zitkovich2023rt}
B.~Zitkovich, T.~Yu, S.~Xu, P.~Xu, T.~Xiao, F.~Xia, J.~Wu, P.~Wohlhart, S.~Welker, A.~Wahid, {\em et~al.}, ``Rt-2: Vision-language-action models transfer web knowledge to robotic control,'' in {\em Conference on Robot Learning}, pp.~2165--2183, PMLR, 2023.

\bibitem{kim2024openvla}
M.~J. Kim, K.~Pertsch, S.~Karamcheti, T.~Xiao, A.~Balakrishna, S.~Nair, R.~Rafailov, E.~Foster, G.~Lam, P.~Sanketi, {\em et~al.}, ``Openvla: An open-source vision-language-action model,'' {\em arXiv preprint arXiv:2406.09246}, 2024.

\bibitem{mees2024octo}
O.~Mees, D.~Ghosh, K.~Pertsch, K.~Black, H.~R. Walke, S.~Dasari, J.~Hejna, T.~Kreiman, C.~Xu, J.~Luo, {\em et~al.}, ``Octo: An open-source generalist robot policy,'' in {\em First Workshop on Vision-Language Models for Navigation and Manipulation at ICRA 2024}, 2024.

\bibitem{ho2020denoising}
J.~Ho, A.~Jain, and P.~Abbeel, ``Denoising diffusion probabilistic models,'' {\em Advances in neural information processing systems}, vol.~33, pp.~6840--6851, 2020.

\bibitem{lipman2022flow}
Y.~Lipman, R.~T. Chen, H.~Ben-Hamu, M.~Nickel, and M.~Le, ``Flow matching for generative modeling,'' {\em arXiv preprint arXiv:2210.02747}, 2022.

\bibitem{black2024pi_0}
K.~Black, N.~Brown, D.~Driess, A.~Esmail, M.~Equi, C.~Finn, N.~Fusai, L.~Groom, K.~Hausman, B.~Ichter, {\em et~al.}, ``pi0: A vision-language-action flow model for general robot control,'' {\em arXiv preprint arXiv:2410.24164}, 2024.

\bibitem{li2024cogact}
Q.~Li, Y.~Liang, Z.~Wang, L.~Luo, X.~Chen, M.~Liao, F.~Wei, Y.~Deng, S.~Xu, Y.~Zhang, {\em et~al.}, ``Cogact: A foundational vision-language-action model for synergizing cognition and action in robotic manipulation,'' {\em arXiv preprint arXiv:2411.19650}, 2024.

\bibitem{reuss2025flower}
M.~Reuss, H.~Zhou, M.~R{\"u}hle, {\"O}.~E. Ya{\u{g}}murlu, F.~Otto, and R.~Lioutikov, ``Flower: Democratizing generalist robot policies with efficient vision-language-action flow policies,'' {\em arXiv preprint arXiv:2509.04996}, 2025.

\bibitem{wang2021structured}
H.~Wang, W.~Wang, W.~Liang, C.~Xiong, and J.~Shen, ``Structured scene memory for vision-language navigation,'' in {\em Proceedings of the IEEE/CVF conference on Computer Vision and Pattern Recognition}, pp.~8455--8464, 2021.

\bibitem{li2025towards}
D.~Li, Y.~Zhang, M.~Cao, D.~Liu, W.~Xie, T.~Hui, L.~Lin, Z.~Xie, and Y.~Li, ``Towards long-horizon vision-language-action system: Reasoning, acting and memory,'' in {\em Proceedings of the IEEE/CVF International Conference on Computer Vision}, pp.~6839--6848, 2025.

\bibitem{shi2025memoryvla}
H.~Shi, B.~Xie, Y.~Liu, L.~Sun, F.~Liu, T.~Wang, E.~Zhou, H.~Fan, X.~Zhang, and G.~Huang, ``Memoryvla: Perceptual-cognitive memory in vision-language-action models for robotic manipulation,'' {\em arXiv preprint arXiv:2508.19236}, 2025.

\bibitem{radford2021learning}
A.~Radford, J.~W. Kim, C.~Hallacy, A.~Ramesh, G.~Goh, S.~Agarwal, G.~Sastry, A.~Askell, P.~Mishkin, J.~Clark, {\em et~al.}, ``Learning transferable visual models from natural language supervision,'' in {\em International conference on machine learning}, pp.~8748--8763, PmLR, 2021.

\bibitem{liu2023visual}
H.~Liu, C.~Li, Q.~Wu, and Y.~J. Lee, ``Visual instruction tuning,'' {\em Advances in neural information processing systems}, vol.~36, pp.~34892--34916, 2023.

\bibitem{steiner2024paligemma}
A.~Steiner, A.~S. Pinto, M.~Tschannen, D.~Keysers, X.~Wang, Y.~Bitton, A.~Gritsenko, M.~Minderer, A.~Sherbondy, S.~Long, {\em et~al.}, ``Paligemma 2: A family of versatile vlms for transfer,'' {\em arXiv preprint arXiv:2412.03555}, 2024.

\bibitem{o2024open}
A.~O’Neill, A.~Rehman, A.~Maddukuri, A.~Gupta, A.~Padalkar, A.~Lee, A.~Pooley, A.~Gupta, A.~Mandlekar, A.~Jain, {\em et~al.}, ``Open x-embodiment: Robotic learning datasets and rt-x models: Open x-embodiment collaboration 0,'' in {\em 2024 IEEE International Conference on Robotics and Automation (ICRA)}, pp.~6892--6903, IEEE, 2024.

\bibitem{ebert2021bridge}
F.~Ebert, Y.~Yang, K.~Schmeckpeper, B.~Bucher, G.~Georgakis, K.~Daniilidis, C.~Finn, and S.~Levine, ``Bridge data: Boosting generalization of robotic skills with cross-domain datasets,'' {\em arXiv preprint arXiv:2109.13396}, 2021.

\bibitem{walke2023bridgedata}
H.~R. Walke, K.~Black, T.~Z. Zhao, Q.~Vuong, C.~Zheng, P.~Hansen-Estruch, A.~W. He, V.~Myers, M.~J. Kim, M.~Du, {\em et~al.}, ``Bridgedata v2: A dataset for robot learning at scale,'' in {\em Conference on Robot Learning}, pp.~1723--1736, PMLR, 2023.

\bibitem{brohan2022rt}
A.~Brohan, N.~Brown, J.~Carbajal, Y.~Chebotar, J.~Dabis, C.~Finn, K.~Gopalakrishnan, K.~Hausman, A.~Herzog, J.~Hsu, {\em et~al.}, ``Rt-1: Robotics transformer for real-world control at scale,'' {\em arXiv preprint arXiv:2212.06817}, 2022.

\bibitem{li2024vision}
X.~Li, M.~Liu, H.~Zhang, C.~Yu, J.~Xu, H.~Wu, C.~Cheang, Y.~Jing, W.~Zhang, H.~Liu, {\em et~al.}, ``Vision-language foundation models as effective robot imitators,'' in {\em International Conference on Learning Representations}, vol.~2024, pp.~26703--26721, 2024.

\bibitem{zheng2025tracevla}
R.~Zheng, Y.~Liang, S.~Huang, J.~Gao, H.~Daum{\'e}~III, A.~Kolobov, F.~Huang, and J.~Yang, ``Tracevla: Visual trace prompting enhances spatial-temporal awareness for generalist robotic policies,'' in {\em International Conference on Learning Representations}, vol.~2025, pp.~54277--54296, 2025.

\bibitem{fan2025interleave}
C.~Fan, X.~Jia, Y.~Sun, Y.~Wang, J.~Wei, Z.~Gong, X.~Zhao, M.~Tomizuka, X.~Yang, J.~Yan, {\em et~al.}, ``Interleave-vla: Enhancing robot manipulation with interleaved image-text instructions,'' {\em arXiv preprint arXiv:2505.02152}, 2025.

\bibitem{bu2025univla}
Q.~Bu, Y.~Yang, J.~Cai, S.~Gao, G.~Ren, M.~Yao, P.~Luo, and H.~Li, ``Univla: Learning to act anywhere with task-centric latent actions,'' {\em arXiv preprint arXiv:2505.06111}, 2025.

\bibitem{gu2023mamba}
A.~Gu and T.~Dao, ``Mamba: Linear-time sequence modeling with selective state spaces,'' {\em arXiv preprint arXiv:2312.00752}, 2023.

\bibitem{dao2024transformers}
T.~Dao and A.~Gu, ``Transformers are ssms: Generalized models and efficient algorithms through structured state space duality,'' {\em arXiv preprint arXiv:2405.21060}, 2024.

\bibitem{tishby2000information}
N.~Tishby, F.~C. Pereira, and W.~Bialek, ``The information bottleneck method,'' {\em arXiv preprint physics/0004057}, 2000.

\bibitem{tishby2015deep}
N.~Tishby and N.~Zaslavsky, ``Deep learning and the information bottleneck principle,'' in {\em 2015 ieee information theory workshop (itw)}, pp.~1--5, Ieee, 2015.

\bibitem{zheng2025universal}
J.~Zheng, J.~Li, D.~Liu, Y.~Zheng, Z.~Wang, Z.~Ou, Y.~Liu, J.~Liu, Y.-Q. Zhang, and X.~Zhan, ``Universal actions for enhanced embodied foundation models,'' in {\em Proceedings of the Computer Vision and Pattern Recognition Conference}, pp.~22508--22519, 2025.

\bibitem{qu2025spatialvla}
D.~Qu, H.~Song, Q.~Chen, Y.~Yao, X.~Ye, Y.~Ding, Z.~Wang, J.~Gu, B.~Zhao, D.~Wang, {\em et~al.}, ``Spatialvla: Exploring spatial representations for visual-language-action model,'' {\em arXiv preprint arXiv:2501.15830}, 2025.

\bibitem{zhao2025cot}
Q.~Zhao, Y.~Lu, M.~J. Kim, Z.~Fu, Z.~Zhang, Y.~Wu, Z.~Li, Q.~Ma, S.~Han, C.~Finn, {\em et~al.}, ``Cot-vla: Visual chain-of-thought reasoning for vision-language-action models,'' in {\em Proceedings of the Computer Vision and Pattern Recognition Conference}, pp.~1702--1713, 2025.

\bibitem{pertsch2025fast}
K.~Pertsch, K.~Stachowicz, B.~Ichter, D.~Driess, S.~Nair, Q.~Vuong, O.~Mees, C.~Finn, and S.~Levine, ``Fast: Efficient action tokenization for vision-language-action models,'' {\em arXiv preprint arXiv:2501.09747}, 2025.

\bibitem{li2025cronusvla}
H.~Li, S.~Yang, Y.~Chen, Y.~Tian, X.~Yang, X.~Chen, H.~Wang, T.~Wang, F.~Zhao, D.~Lin, {\em et~al.}, ``Cronusvla: Transferring latent motion across time for multi-frame prediction in manipulation,'' {\em arXiv e-prints}, pp.~arXiv--2506, 2025.

\bibitem{liu2025trivla}
Z.~Liu, Y.~Gu, S.~Zheng, Y.~Fu, X.~Xue, and Y.-G. Jiang, ``Trivla: A triple-system-based unified vision-language-action model with episodic world modeling for general robot control,'' {\em arXiv preprint arXiv:2507.01424}, 2025.

\bibitem{zhang20264d}
J.~Zhang, Y.~Chen, Y.~Xu, Z.~Huang, Y.~Zhou, Y.-J. Yuan, X.~Cai, G.~Huang, X.~Quan, H.~Xu, {\em et~al.}, ``4d-vla: Spatiotemporal vision-language-action pretraining with cross-scene calibration,'' {\em Advances in Neural Information Processing Systems}, vol.~38, pp.~33914--33937, 2026.

\bibitem{intelligence2025pi_}
P.~Intelligence, K.~Black, N.~Brown, J.~Darpinian, K.~Dhabalia, D.~Driess, A.~Esmail, M.~Equi, C.~Finn, N.~Fusai, {\em et~al.}, ``pi0.5: a vision-language-action model with open-world generalization,'' {\em arXiv preprint arXiv:2504.16054}, 2025.

\bibitem{liu2023libero}
B.~Liu, Y.~Zhu, C.~Gao, Y.~Feng, Q.~Liu, Y.~Zhu, and P.~Stone, ``Libero: Benchmarking knowledge transfer for lifelong robot learning,'' {\em Advances in Neural Information Processing Systems}, vol.~36, pp.~44776--44791, 2023.

\bibitem{mees2022calvin}
O.~Mees, L.~Hermann, E.~Rosete-Beas, and W.~Burgard, ``Calvin: A benchmark for language-conditioned policy learning for long-horizon robot manipulation tasks,'' {\em IEEE Robotics and Automation Letters}, vol.~7, no.~3, pp.~7327--7334, 2022.

\bibitem{reuss2024multimodal}
M.~Reuss, {\"O}.~E. Ya{\u{g}}murlu, F.~Wenzel, and R.~Lioutikov, ``Multimodal diffusion transformer: Learning versatile behavior from multimodal goals,'' {\em arXiv preprint arXiv:2407.05996}, 2024.

\bibitem{liu2024robouniview}
F.~Liu, F.~Yan, L.~Zheng, C.~Feng, Y.~Huang, and L.~Ma, ``Robouniview: Visual-language model with unified view representation for robotic manipulation,'' {\em arXiv preprint arXiv:2406.18977}, 2024.

\bibitem{wang2026think}
F.~Wang, P.~Zhou, J.~Qi, S.~Lyu, D.~Navarro-Alarcon, and G.~Guo, ``Think proprioceptively: Embodied visual reasoning for vla manipulation,'' {\em arXiv preprint arXiv:2602.06575}, 2026.

\bibitem{zhou2024towards}
P.~Zhou, X.~Xie, Z.~Lin, and S.~Yan, ``Towards understanding convergence and generalization of adamw,'' {\em IEEE transactions on pattern analysis and machine intelligence}, vol.~46, no.~9, pp.~6486--6493, 2024.

\bibitem{beyer2024paligemma}
L.~Beyer, A.~Steiner, A.~S. Pinto, A.~Kolesnikov, X.~Wang, D.~Salz, M.~Neumann, I.~Alabdulmohsin, M.~Tschannen, E.~Bugliarello, {\em et~al.}, ``Paligemma: A versatile 3b vlm for transfer,'' {\em arXiv preprint arXiv:2407.07726}, 2024.

\bibitem{xiao2024florence}
B.~Xiao, H.~Wu, W.~Xu, X.~Dai, H.~Hu, Y.~Lu, M.~Zeng, C.~Liu, and L.~Yuan, ``Florence-2: Advancing a unified representation for a variety of vision tasks,'' in {\em Proceedings of the IEEE/CVF Conference on Computer Vision and Pattern Recognition}, pp.~4818--4829, 2024.

\end{thebibliography}

\clearpage
\appendix

\begin{table}
\centering
\begin{tabular}{lc}
\toprule
\textbf{Module} & \textbf{Number of Parameters} \\
\midrule
SSM   & 432.14 K  \\
Slot Pooling  & 2.06 K  \\
VIC  & 131.58 K \\
Memory Fusion \& Gate  & 2.10 M  \\
Projection & 787.71 K  \\
\midrule
\textbf{TOTAL} & \textbf{3.45 M} \\
\bottomrule
\end{tabular}
\caption{Parameter counts of different components in \textbf{CogMemory}.}
\label{tab:param}
\end{table}

\begin{table}
\centering
\begin{tabular}{lc}
\toprule
\textbf{Hyperparameters} & \textbf{Values} \\
\midrule
GPU type   & NVIDIA L20X \\
GPU num  & 8  \\
fsdp devices & 8 \\
\midrule
Optimizer & Adamw \\
- b1 & 0.9 \\
- b2 & 0.95 \\
- eps & 1e-8 \\
- weight decay & 1e-10 \\
- clip gradient norm & 1.0\\
\midrule
lr schedule & CosineDecay\\
- warmup steps & 1k \\
- peak lr & 5e-5 \\
- decay steps & 30k \\
- final lr & 5e-6 \\
\midrule
ema decay & 0.999 \\ 
Random seed & 42 \\
Batch size & 256 \\
Data loader workers & 2 \\
$\log\sigma_k^2$ clip range & [-6.0, 2.0] \\
$\lambda$ of KL loss& 1e-4 \\
Action dim & 7 \\
Action horizon & 10 \\
Total steps & 30k \\
\bottomrule
\end{tabular}
\caption{Hyperparameters for training \textbf{CogMemory}.}
\label{tab:hyperparam}
\end{table}

\section{TRAINING DETAILS}

\subsection{Model Complexity Analysis}

Table~\ref{tab:param} summarizes the parameter counts of different components in \textbf{CogMemory}. The complete memory module introduces only 3.45M additional parameters, demonstrating the lightweight design of the proposed framework. Among all components, the memory fusion and gating module contributes the largest portion of parameters due to the cross-attention based memory integration layers, while the temporal abstraction and variational compression modules remain highly compact. Despite the small parameter overhead, \textbf{CogMemory} consistently improves long-horizon manipulation performance across multiple VLA backbones and benchmarks.

\begin{algorithm}
\caption{\textbf{CogMemory} for Long-Horizon VLA Policies}
\label{alg:cogmemory}
\KwIn{
Multimodal representations $X=[x_1,\dots,x_L]$,
memory slot number $K$,
fusion type $\mathcal{F}\in\{\texttt{early},\texttt{intermediate}\}$
}
\KwOut{Optimized policy with actions $\hat{a}_{1:T}$}

\BlankLine
\textbf{// Stage I: Cognitive Memory Abstraction}

Encode temporal history using selective SSM:
$H=\mathrm{SSM}(X)$\;

Compute slot assignments:
$A=HW_p$ and
$\alpha_{:,k}=\mathrm{Softmax}_{t}(A_{:,k})$\;

\For{$k=1$ \KwTo $K$}{
Aggregate temporal memory slot:
$m_k=\sum_t \alpha_{t,k}h_t$\;

Estimate latent distribution:
$\mu_k=f_\mu(m_k)$,
$\log\sigma_k^2=f_\sigma(m_k)$\;

Sample latent memory:
$z_k=\mu_k+\sigma_k\odot\epsilon$,
$\epsilon\sim\mathcal{N}(0,I)$\;
}

Project latent variables into cognitive memory:
$M=f_{\mathrm{proj}}(Z)$\;

Compute KL regularization:
$\mathcal{L}_{KL}=\sum_k D_{KL}(q(z_k|m_k)\|p(z_k))$\;

\BlankLine
\textbf{// Stage II: Memory-Guided Action Fusion}

\eIf{$\mathcal{F}=\texttt{early}$}{

Extract action suffix representations:
$Q_{\mathrm{suffix}}\subset X$\;

Retrieve task-relevant memory:
$R=\mathrm{Softmax}(Q_{\mathrm{suffix}}M^\top/\sqrt{D})M$\;

Adaptive gated fusion:
$\tilde{Q}=Q_{\mathrm{suffix}}
+\sigma(W_g[Q_{\mathrm{suffix}};R])\odot R$\;

Predict actions:
$\hat{a}_{1:T}=\mathrm{Decoder}(\tilde{Q})$\;

}{

\ForEach{Flow Transformer layer $l$}{

Extract intermediate hidden states:
$Q^{(l)}$\;

Retrieve memory:
$R^{(l)}=\mathrm{Softmax}(Q^{(l)}M^\top/\sqrt{D})M$\;

Inject cognitive memory:
$\tilde{Q}^{(l)}
=
Q^{(l)}
+
\sigma(W_g[Q^{(l)};R^{(l)}])\odot R^{(l)}$\;
}

Predict actions from final hidden states\;
}

Optimize:
$\mathcal{L}
=
\mathcal{L}_{\mathrm{action}}
+
\lambda\mathcal{L}_{KL}$\;

\end{algorithm}

\subsection{Training Hyperparameters}

Table~\ref{tab:hyperparam} presents the detailed training hyperparameters used in all experiments. Unless otherwise specified, all models are trained using AdamW optimization with cosine learning-rate decay and exponential moving average (EMA) updates. We train \textbf{CogMemory} using 8 NVIDIA L20X GPUs with a global batch size of 256. The variational memory module uses KL regularization with coefficient $\lambda=10^{-4}$ and clips the latent log-variance within $[-6.0, 2.0]$ for stable training. All experiments use an action horizon of 10 and are trained for 30k optimization steps. We train \textbf{CogMemory} on two representative VLA backbones with different fusion paradigms: the PaliGemma \cite{beyer2024paligemma} backbone ($\pi_{0.5}$-style, 3B parameters) for early-fusion architectures and the Florence-2-L \cite{xiao2024florence} backbone (FLOWER-style, 0.77B parameters) for intermediate-fusion architectures.

\subsection{CogMemory Algorithm}

Algorithm~\ref{alg:cogmemory} summarizes the overall training and inference procedure of \textbf{CogMemory}. The framework first constructs compact cognitive memory representations through selective state-space temporal encoding, multi-slot temporal abstraction, and variational latent compression. During policy inference, the memory representations are dynamically integrated into downstream action generation through memory-conditioned retrieval and adaptive gated fusion. The framework supports both early-fusion and intermediate-fusion VLA architectures within a unified training formulation.

\section{DATASETS DETAILS.}

\begin{table}
\centering
\small
\begin{tabular}{lc}
\toprule
\textbf{Property} & \textbf{Value} \\
\midrule
Robot Platform & Franka Panda \\
Simulation Engine & MuJoCo \\
Observation FPS & 10 \\
Top Camera Resolution & $256\times256$ \\
Wrist Camera Resolution & $256\times256$ \\
Robot State Dimension & 8 \\
Action Dimension & 7 \\
\midrule
Spatial / Object / Goal / Long-10 & 40 tasks total \\
Training Episodes (40 tasks) & 1,693 \\
Training Frames (40 tasks) & 273,465 \\
\midrule
Long-90 Tasks & 90 tasks \\
Training Episodes (Long-90) & 4500 \\
Training Frames (Long-90) & 669043 \\
\midrule
Demonstrations per Task & 50 \\
Evaluation Trails & 50 \\
\bottomrule
\end{tabular}
\caption{Statistics of the LIBERO benchmark used in our experiments.}
\label{tab:libero_stats}
\end{table}

\begin{table}[t]
\centering
\small
\begin{tabular}{lc}
\toprule
\textbf{Property} & \textbf{Value} \\
\midrule
Robot Platform & Franka Emika\\
Simulation Engine & PyBullet \\
Atomic Manipulation Skills & 34 \\
Language Instructions & 389 annotations \\
Evaluation Protocol & 1000 instruction chains \\
Instructions per Chain & 5 tasks \\
Observation FPS & 10 \\
Top Camera Resolution & $200\times200$ \\
Wrist Camera Resolution & $84\times84$ \\
Robot State Dimension & 15 \\
Action Dimension & 7 \\
\midrule
ABC Training Episodes & 18,957 \\
ABC Training Frames & 1,136,536 \\
ABC Training Split & Environments A/B/C \\
ABC Testing Split & Environment D \\
\midrule
D Training Episodes & 6,135 \\
D Training Frames & 369,493 \\
D Training / Testing & Environment D \\
\bottomrule
\end{tabular}
\caption{Statistics of the CALVIN benchmark used in our experiments.}
\label{tab:calvin_stats}
\end{table}

\subsection{LIBERO Benchmark}

We evaluate \textbf{CogMemory} on the LIBERO benchmark~ \cite{liu2023libero}, a large-scale benchmark designed for evaluating long-horizon and language-conditioned robotic manipulation. LIBERO is built upon the BDDL task specification framework and focuses on compositional generalization across diverse household manipulation scenarios. The benchmark is implemented in the MuJoCo simulator using a Franka Panda robot equipped with a parallel gripper.

LIBERO consists of five evaluation suites with increasing levels of difficulty and compositional complexity: \textit{LIBERO-Spatial}, \textit{LIBERO-Object}, \textit{LIBERO-Goal}, \textit{LIBERO-Long-10}, and \textit{LIBERO-Long-90}. The Spatial, Object, Goal, and Long-10 suites each contain 10 tasks, while Long-90 contains 90 long-horizon tasks. The benchmark includes a diverse set of tabletop manipulation behaviors such as object picking, placing, opening, closing, rearrangement, drawer interaction, and multi-stage task execution.

The Spatial suite evaluates generalization across different object positions and scene layouts, requiring policies to handle spatial variation while preserving task semantics. The Object suite focuses on unseen object combinations and evaluates object-level compositional generalization. The Goal suite further increases difficulty by introducing diverse goal configurations and longer manipulation sequences. In contrast, the Long-10 and Long-90 suites emphasize persistent temporal reasoning and long-horizon sequential manipulation, where policies must maintain consistent task progress across multiple interaction stages.

Each task is paired with natural language instructions describing the manipulation objective. Following the standard LIBERO protocol, demonstrations are collected using scripted policies and teleoperation trajectories. The benchmark provides RGB observations, proprioceptive robot states, and language annotations for policy training. Table~\ref{tab:libero_stats} presents the detailed statistical information of the dataset that we actually used.

\subsection{CALVIN Benchmark}

We further evaluate \textbf{CogMemory} on the CALVIN benchmark~ \cite{mees2022calvin}, a widely used benchmark for long-horizon language-conditioned robotic manipulation. CALVIN is designed to evaluate sequential decision making, temporal consistency, and generalization under multi-step instruction chaining. The benchmark is built in a simulated tabletop household environment using a Franka Emika robot equipped with RGB cameras and proprioceptive sensing.

CALVIN contains a large collection of manipulation trajectories generated from diverse interaction primitives, including opening and closing drawers, sliding doors, rotating switches, pressing buttons, moving blocks, and object relocation. The environment includes multiple articulated objects and interactive scene elements, resulting in complex long-horizon task dependencies. Policies receive RGB observations, robot proprioceptive states, and free-form language instructions as input.

A key characteristic of CALVIN is its long-horizon evaluation protocol. Instead of evaluating single-task execution, policies are required to sequentially complete multiple language instructions within a single rollout episode. Evaluation is performed over 1,000 instruction chains, where each chain contains five consecutive manipulation tasks sampled from the benchmark task set. A policy succeeds at chain length LH-$n$ if it correctly completes the first $n$ tasks in the instruction sequence without failure. The benchmark additionally reports the average completed chain length (Avg. Len.), which measures long-horizon task consistency and persistent temporal reasoning capability.

CALVIN provides two standard evaluation settings: ABC$\rightarrow$D and D$\rightarrow$D. In the ABC$\rightarrow$D setting, policies are trained on environments A/B/C and evaluated on the unseen environment D, emphasizing cross-environment generalization and zero-shot transfer capability. In contrast, D$\rightarrow$D evaluates policies within the same environment distribution, focusing primarily on long-horizon sequential manipulation performance. The ABC$\rightarrow$D protocol is generally considered substantially more challenging due to distribution shift in scene layouts, object appearances, and environmental configurations. Table~\ref{tab:calvin_stats} presents the detailed statistical information of the dataset that we actually used.

\section{INFERENCE EFFICIENCY ANALYSIS}

To further evaluate the practical efficiency of \textbf{CogMemory}, we compare the execution time between the baseline policy and \textbf{CogMemory} on a set of successfully completed manipulation tasks. The evaluated tasks cover diverse manipulation primitives, including picking, placing, stacking, opening, and closing operations.

As summarized in Table~\ref{tab:efficiency}, we report the execution time of tasks that are successfully completed by both methods, enabling a fair comparison of execution efficiency under successful task completion. \textbf{CogMemory} achieves a lower average execution time of 12.9s compared with 21.2s for the baseline policy. These results indicate that compact memory-guided action generation improves execution efficiency by enabling more consistent and effective long-horizon decision making without introducing additional execution overhead.

\section{ADDITIONAL QUALITATIVE RESULTS}

We provide additional qualitative results of \textbf{CogMemory} on both LIBERO and CALVIN benchmarks. The visualizations illustrate representative long-horizon manipulation trajectories under different task settings, demonstrating the ability of \textbf{CogMemory} to preserve task-relevant historical context and maintain stable sequential decision making over extended interaction horizons.

\begin{figure*}[t]
\centering
\includegraphics[width=0.99\textwidth]{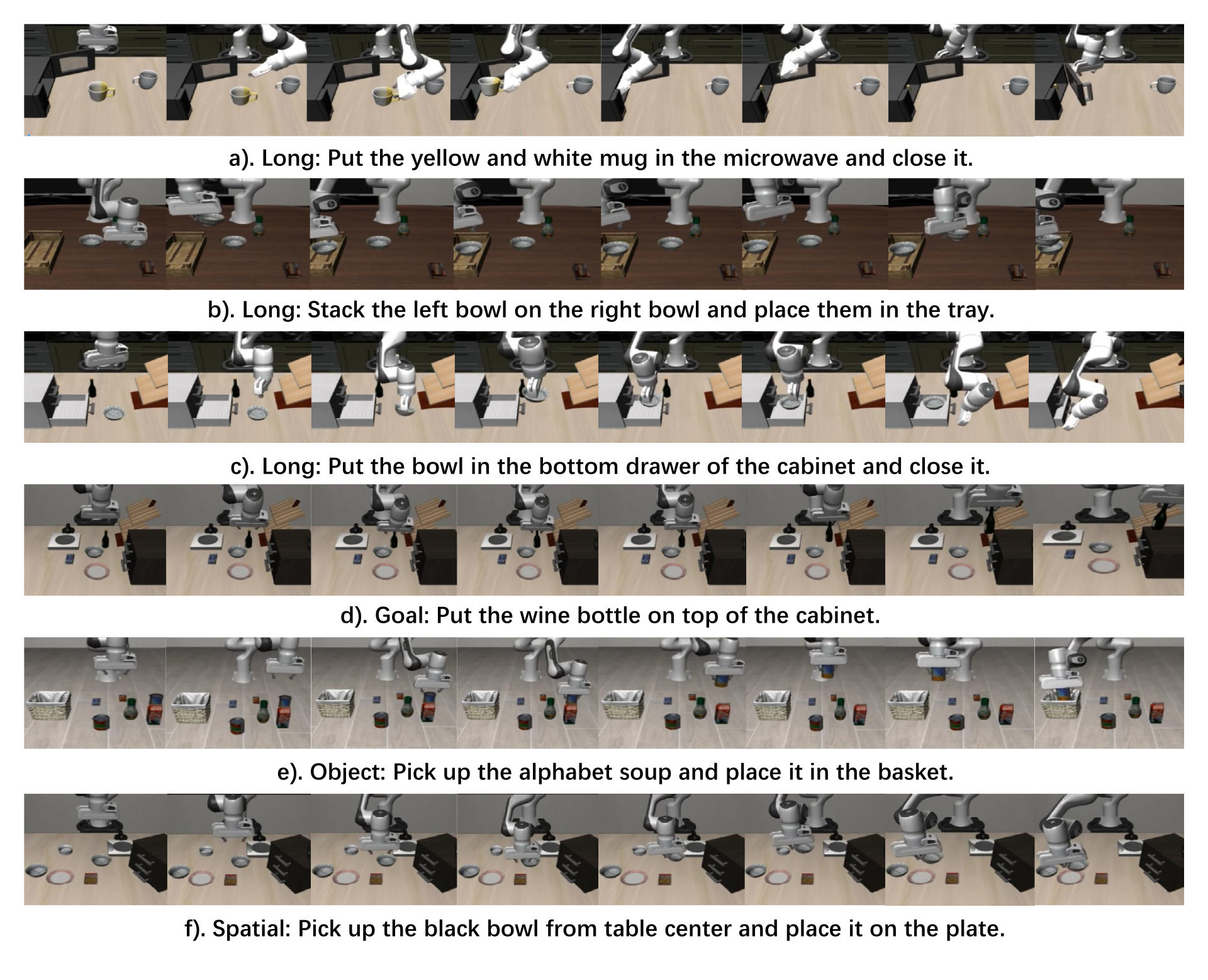}
\caption{\textbf{Qualitative results of CogMemory on Libero tasks.} Representative examples include tasks from Spatial, Object, Goal, Long-10, and Long-90 suites.}
\label{fig:libero}
\end{figure*}

\begin{figure*}
\centering
\includegraphics[width=\textwidth]{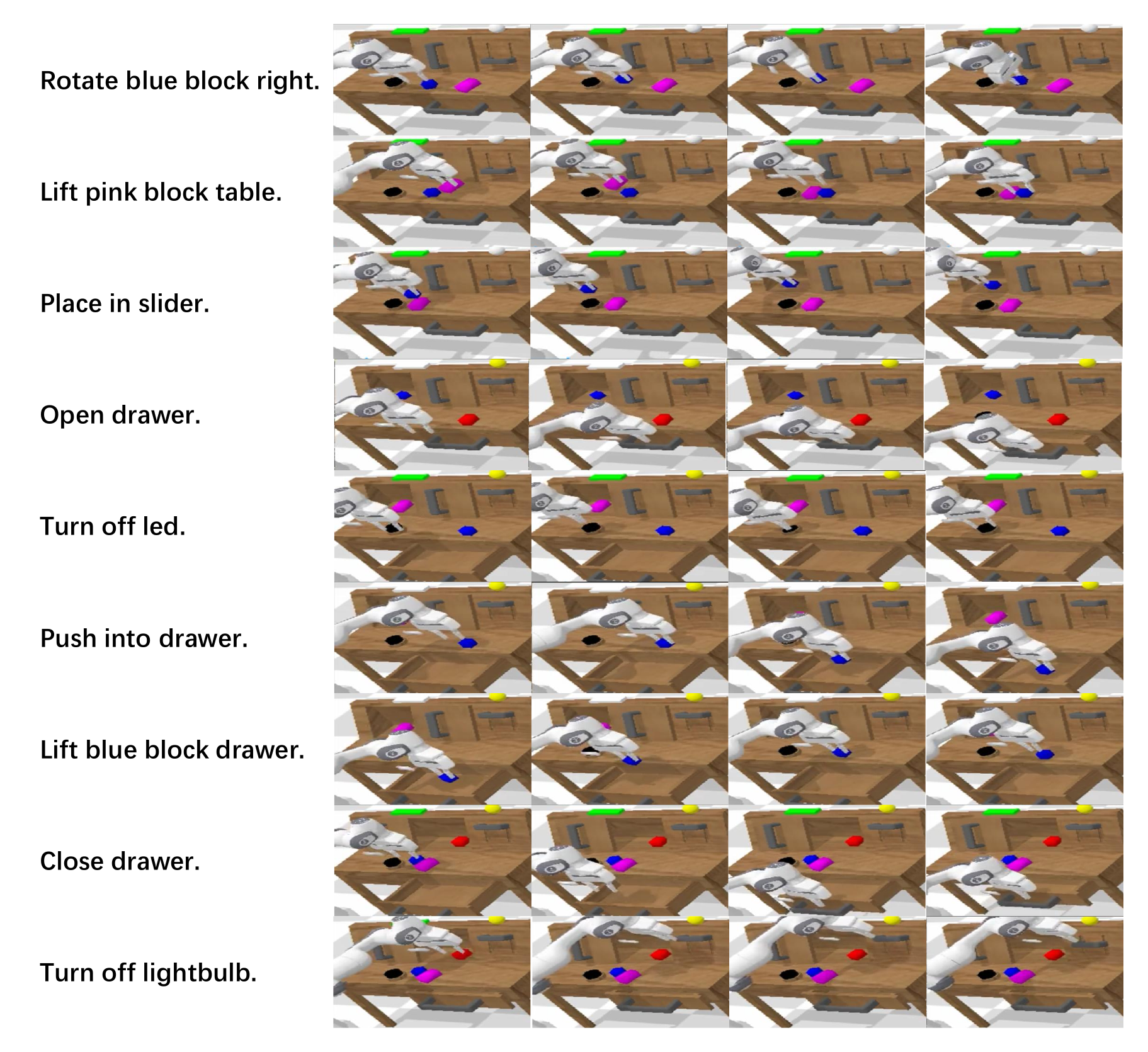}
\caption{\textbf{Qualitative results of CogMemory on Calvin tasks.} Representative examples include tasks from Calvin validation set.}
\label{fig:calvin}
\end{figure*}

Figure~\ref{fig:libero} presents qualitative examples on LIBERO across multiple task suites, including Spatial, Object, Goal, Long-10, and Long-90. Figure~\ref{fig:calvin} shows representative long-horizon manipulation sequences on the CALVIN benchmark.

\begin{table}
\centering
\small
\renewcommand{\arraystretch}{0.85}
\begin{tabular}{p{0.5\linewidth}cc}
\toprule
\textbf{Task} & \textbf{Baseline} & \textbf{CogMemory} \\
\midrule
Close bottom drawer & 0:24 & 0:09 \\
Close microwave & 0:30 & 0:17 \\
Close top drawer & 0:20 & 0:06 \\
Open bottom drawer & 0:15 & 0:14 \\
Open top drawer & 0:35 & 0:08 \\
Soup $\rightarrow$ basket & 0:22 & 0:11 \\
Soup $\rightarrow$ tray & 0:11 & 0:10 \\
Left black bowl $\rightarrow$ tray & 0:10 & 0:10 \\
Book $\rightarrow$ back caddy & 0:15 & 0:14 \\
Book $\rightarrow$ left caddy & 0:21 & 0:13 \\
Butter $\rightarrow$ basket & 0:26 & 0:10 \\
Butter $\rightarrow$ tray & 0:26 & 0:08 \\
Chocolate pudding $\rightarrow$ tray & 0:18 & 0:13 \\
Cream cheese $\rightarrow$ tray & 0:15 & 0:15 \\
Cream cheese box $\rightarrow$ basket & 0:31 & 0:15 \\
Ketchup $\rightarrow$ basket & 0:23 & 0:16 \\
Orange juice $\rightarrow$ basket & 0:20 & 0:13 \\
Tomato sauce $\rightarrow$ basket & 0:27 & 0:09 \\
Tomato sauce $\rightarrow$ tray & 0:19 & 0:10 \\
Black bowl (back) $\rightarrow$ plate & 0:09 & 0:09 \\
Black bowl $\rightarrow$ bottom drawer & 0:14 & 0:10 \\
Black bowl $\rightarrow$ top drawer & 0:12 & 0:11 \\
Black bowl $\rightarrow$ plate & 0:35 & 0:10 \\
Black bowl $\rightarrow$ cabinet top & 0:26 & 0:14 \\
Chocolate pudding $\rightarrow$ right plate & 0:22 & 0:07 \\
Frying pan $\rightarrow$ cabinet top & 0:30 & 0:21 \\
Middle black bowl $\rightarrow$ plate & 0:10 & 0:09 \\
Moka pot $\rightarrow$ stove & 0:36 & 0:14 \\
Right moka pot $\rightarrow$ stove & 0:37 & 0:21 \\
White bowl $\rightarrow$ plate & 0:31 & 0:17 \\
White bowl $\rightarrow$ cabinet top & 0:13 & 0:12 \\
White mug $\rightarrow$ left plate & 0:10 & 0:09 \\
White mug $\rightarrow$ plate & 0:11 & 0:15 \\
Wine bottle $\rightarrow$ bottom drawer & 0:14 & 0:12 \\
Yellow-white mug $\rightarrow$ right plate & 0:22 & 0:08 \\
Stack bowls $\rightarrow$ tray & 0:36 & 0:18 \\
Turn on stove & 0:09 & 0:09 \\
\bottomrule
\end{tabular}
\caption{Execution time comparison between the baseline policy and \textbf{CogMemory} on tasks successfully completed by both methods. Task names are abbreviated for readability.}
\label{tab:efficiency}
\end{table}

\clearpage

\end{document}